\documentclass[letterpaper, 10 pt, conference]{ieeeconf}

\IEEEoverridecommandlockouts 
\title{\LARGE \bf
ECNNs: Ensemble Learning Methods for Improving Planar Grasp Quality Estimation
}

\author{Fadi Alladkani$^{1}$, James Akl$^{1}$, Berk Calli$^{1}$
\thanks{$^{1}$Authors are with Robotics Engineering Program, Worcester Polytechnic Institute, 85 Prescott Street, Worcester, MA-01605, USA
        {\tt\small \{fmalladkani;jgakl;bcalli\}@wpi.edu}
        This work was supported in part by the National Science Foundation under grant FW-HTF-1928506.
        }
}
\usepackage{soul} 
\usepackage[vlined,ruled]{algorithm2e}
\SetKwProg{Fn}{Function}{}{}
\usepackage{cite}
\usepackage{amsmath,amssymb,amsfonts}
\usepackage{algorithmic}
\usepackage{graphicx}
\usepackage{textcomp}
\usepackage{xcolor}
\usepackage{gensymb}
\usepackage[utf8]{inputenc}
\usepackage{array}
\usepackage{wrapfig}
\usepackage{multirow}
\usepackage{tabu}
\usepackage{fullpage}
\usepackage{times}
\let\labelindent\relax 
\usepackage{enumitem,kantlipsum}
\usepackage{fancyhdr,graphicx,amsmath,amssymb}
\usepackage[ruled,vlined]{algorithm2e}
\include{pythonlisting}
\usepackage{lipsum}
\usepackage{multicol}
\usepackage{hyperref}

\begin{document}

\maketitle
\thispagestyle{empty}
\pagestyle{empty}

\begin{abstract}
We present an ensemble learning methodology that combines multiple existing robotic grasp synthesis algorithms and obtain a success rate that is significantly better than the individual algorithms. The methodology treats the grasping algorithms as ``experts" providing grasp ``opinions". An Ensemble Convolutional Neural Network (ECNN) is trained using a Mixture of Experts (MOE) model that integrates these opinions and determines the final grasping decision.  The ECNN introduces minimal computational cost overhead, and the network can virtually run as fast as the slowest expert. We test this architecture using open-source algorithms in the literature by adopting GQCNN 4.0, GGCNN and a custom variation of GGCNN as experts and obtained a 6\% increase in the grasp success on the Cornell Dataset compared to the best-performing individual algorithm. The performance of the method is also demonstrated using a Franka Emika Panda arm.

\end{abstract}

\section{INTRODUCTION}

Many service robotics and warehouse applications require to grasp a large variety of novel objects, whose models are not available to the robotic system a priori. There has been major improvements in the robotic grasping domain in the last decade \cite{dexnet2, generative, newGenerative, earlyGrasping, multigrasp}, many of which are machine learning-based frameworks providing a grasping policy that can generalize over previously unseen objects. Unfortunately, the potential set of inputs to these algorithms is incredibly large and the variation in objects to grasp, lighting, and camera conditions, can cause losses in performance and accuracy.  Many algorithms have been trained and are capable of mapping a quite significant subset of this input space, but still lose reliability when encountering inputs outside of this space. \cite{DifficultiesInBenchmarking} showcases some of the challenges in reproducing results due to the large variation in conditions. This issue is inherent in the architectures of the neural networks, and is compounded here compared to other fields (e.g. computer vision) due to the more complex and variable nature of the inputs involved in robotic grasping.

\begin{figure}
    \includegraphics[width=3.0in]{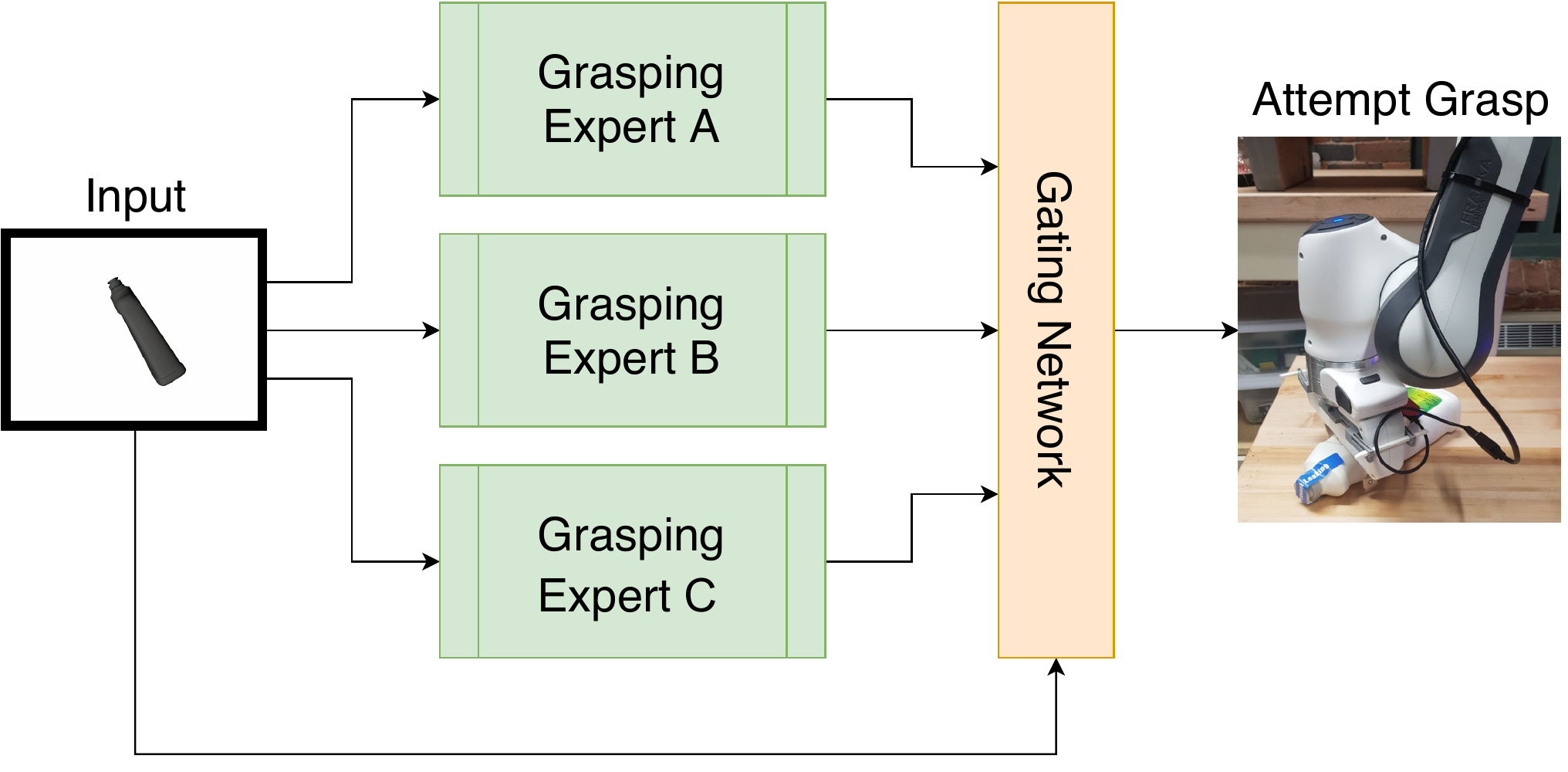}
    \caption{The proposed ensemble learning-based strategy utilizes the opinions of grasping experts and achieves a better accuracy compared to individual experts themselves.  The gating network is a neural network trained to assign a set of weights to each expert based on the input data.}
    \label{fig:figure 1}
\end{figure}

Instead of aiming to develop a silver bullet grasp synthesis algorithm that can generalize over any possible object and condition, our research seeks to integrate the outputs of multiple grasping algorithms to achieve a higher degree of robustness and accuracy compared to the individual algorithms themselves. We achieve this goal via an ensemble learning framework \cite{ensembleSurvey} in which the individual grasp synthesis algorithms are treated as ``experts" providing their ``opinion" for a given input, and a supervisory system trained using convolutional neural networks intelligently combines these opinions and outputs the final decision (Fig.~\ref{fig:figure 1}). We implement this framework by developing a Mixture of Experts (MOE) ensemble model by adopting GQCNN-4.0 \cite{gqcnn_latest}, GGCNN \cite{generative} and GGCNN-RGB (our custom implementation of GGCNN using RGB images) as experts. We run our MOE model on the Cornell Grasp Dataset \cite{planarGrasp1}, and compared its performance to the individual success rates of the experts. We obtained a 6\% improvement over the best performing individual network, which is achieved via the ability of the MOE model to combine the advantages of each expert, while overcoming their individual disadvantages through learning how to best partition the input space. We also provide experiments using a Franka Emika robot and obtained significant improvements via our ensemble strategy compared to individual expert opinions.

This strategy allows us to take advantage of multiple grasp synthesis approaches and therefore provide better generalization overall. The methodology introduces minimal computational cost overhead, allowing calculations to be as fast as the slowest expert. We present efficient means to train the network that avoids recalculations by caching expert results.

\section{Related Work}\label{sec: litreview}
In this section, we briefly summarize the grasp synthesis literature and provide a background for the ensemble learning framework.
\subsection{Grasp Synthesis Literature}
Convolutional Neural Networks dominate the robotic grasp synthesis domain \cite{dataDrivenSurvey}. The most common methods, such as those in \cite{planarGrasp1, planarGrasp2, dexnet2} rely on a common pipeline of grasp candidate sampling followed by grasp quality ranking. This pipeline is adopted here to enable the combination of multiple grasping algorithms in the quality ranking stage. Recent approaches, notably \cite{generative} and \cite{newGenerative} have attempted to improve the performance of such techniques by relying on generative methods, unifying both the sampling and ranking steps into one, improving performance significantly, and enabling fast closed-loop grasping, and such generative algorithms have been incorporated as an expert in our ensemble network. \cite{multigrasp} improves on typical approaches by performing multi-object, multi-grasp classification in one shot by defining the learning problem as a null hypothesis competition. It is of important note that many of these algorithms are open-source, and our paper utilizes the libraries provided by \cite{gqcnn_latest, generative}.

\subsection{Ensemble Learning Literature}
Ensemble-based learning framework presents a class of techniques that take advantage of multiple individual algorithms having a difference in opinion. The framework aims to improve the performance of the system as a whole through intelligent voting-based procedures.  These techniques include popular methods such as boosting \cite{boosting} and bagging \cite{bagging}, well-established in the literature as effective performance boosters for improving accuracy (\cite{ensembleSurvey}). A more advanced method of ensemble learning that is more flexible and adaptable to separate algorithms is the MOE  model \cite{original_moe}.  MOE-based frameworks have been used extensively throughout multiple fields \cite{MOEHistory}, and have been known to be top performers in several competitions (\cite{ASC}).

MOEs take in experts that have a difference in opinion, and attempts to combine their output intelligently using a gating network which learns how to best assign weights to the experts.  This combination can happen in several different ways:  \cite{fundus} utilize the simplest form of this method, which averages over the outputs of the individual networks. This approach works best for directly compatible outputs and inputs, where averaging produces sensible results. This is unsuitable for our purposes since it requires the outputs to be in the same form. This limits the choice of available experts. Thus a more flexible combination approach is desired. \cite{ASC} performs a weighted summation by using a neuron connected to the outputs of all experts, enabling it to learn the best weights to assign, allowing for a more performant combination of outputs compared to simple averaging.  The MOE system, however, can contain weights that vary based on the input being fed into the experts, which is the primary approach adopted in this paper. For such a system, the gating network may itself be a separate neural network that takes in the input, and provides the set of weights to be assigned to each expert, such as the one presented in \cite{MOEMovie}, which relies on the full MOE model to perform emotion prediction in movies.  In our paper, we present two different methods for performing input-dependant weight assignment to the grasping experts i.e. the image of the target object(s) that is used for the experts to determine grasp opinions is also used to determine the weights of the ensemble network. For different objects/images, the MOE provides different weights. There have been several interesting uses for ensemble-based learning in robotics. \cite{reinforcementSensorimotor} uses an MOE model with completely heterogeneous learners that enable the robot to learn sensorimotor representations, providing an important variety in opinions.  \cite{randomCroppingEnsemble} uses an ensemble-based framework for better cropping of images in highly cluttered and unevenly-illuminated scenes, enabling a grasping algorithm to then attempt to pick the cropped piece, demonstrating its ability to adapt to highly complex and variable scenes, but such an approach does not focus on boosting network performance, but on image and object isolation.  \cite{rotationalInvarianceEnsemble} uses an ensemble-based framework to achieve rotational invariance in fully convolutional neural networks, which can boost grasping CNN performance. The closest paper to our current approach is \cite{asif2018ensemblenet}, which utilizes an ensemble gating network to assign weights to the grasps generated by four heterogeneous experts, and the expert with the maximum assigned weight is chosen.  This approach is similar to our use of an ensemble to improve overall accuracy.  However, our approach focuses not on expert output selection, but on expert output combination, a more flexible method that combines the information from multiple experts to best estimate the grasp quality.

\section{Problem Formulation}
\label{sec:problem_formulation}
Here we discuss the representations of a grasp on an image, along with the definition of the role of a grasping expert. Then, we formulate the MOE model and define all the necessary variables, methods, and equations involved.  We also make explicit the requirements and constraints necessary to make the MOE model work.

\subsection{Grasping}
\label{sec:grasp_formulation}


Continuing in the vein of the recent grasping literature, we tackle the problem of associating planar, antipodal grasps on a given n-channel image to a certain quality value indicative of its odds of success.

Let an image of the object be $\mathbf{I}=\mathbb{R}^{W \times H \times N}$, which is an image with width $W$, height $H$, and channels $N$. Here $W$ and $H$ refer to the number of pixels.  An antipodal grasp can be defined on $\mathbf{I}$ by

\begin{equation}
    \mathbf{G}_\mathrm{img}=(u, v, d, w, \theta, q)
\end{equation}

The location of this grasp on the image is defined by its position $(u, v, d)$, where $u$ and $v$ are its pixel coordinates along the horizontal and vertical image axes respectively, and $d$ is the depth; its width $w$; and its angle clockwise with the horizontal $\theta$ with a quality $q$, which is indicative of its stability.  We define the grasp in the world frame to be $\mathbf{G}$, which can be obtained by the mapping $\mathbf{G} = f(\mathbf{G}_\mathrm{img})$, which is a function of the camera intrinsics, extrinsics, and robot pose.  Once transformed into the world frame, the robot may proceed to execute the resulting grasp.  We also define the set of all grasps on the image as $\mathbf{G}$.

The goal of a grasping algorithm is to approximate the quality of a given grasp.  Setting the approximated grasp quality as $\Tilde{q}$, a grasping algorithm, or expert, can be defined as one that maps a grasp location $(u, v, d, w, \theta)$ on an image, into an estimated quality

\begin{equation}
    \Tilde{q} = y(u, v, d, w, \theta)
\end{equation}

In the presence of multiple experts, they are indexed as $y_i$, with the total number of experts being $n$.

\subsection{Mixture of Experts}
\label{sec:moe_formulation}
A mixture of experts model is a type of ensemble-based model.  Ensemble models make use of multiple individual learners, called experts, which are either heterogeneous (two or more different expert types) or homogeneous (two or more of the same expert type) \cite{ensembleTypes} and attempt to combine their output intelligently such that the accuracy of the system as a whole improves.


The combination of the usable expert outputs $y_i$ can happen in multiple different ways.  The goal of an MOE model is to determine the optimal values for the weights $g_i(x)$ given to each expert such that the output of the full system $y$ is as close to ground truth as possible.

\begin{equation}
    y(x) = \sum_{i=1}^{n}{y_i(x)\,g_i(x)}
\end{equation}

We illustrate three methods for which the values for $g_i(x)$ can be calculated, in increasing order of complexity:

\begin{enumerate}[leftmargin=*]
    \item Simple averaging (${g_i(x)=1/n}$).  A method such as this mainly works when the output of each network is simple enough.
    \item A linear combination of the experts ($g_i(x)=g_i$).  This can best be seen as a single fully connected neuron with inputs being the outputs of the experts.
    \item A linear combination of the experts dependant on the input.  This is the full MOE formulation that requires a separate gating network to estimate $g_i(x)$.  A softmax layer at the output can ensure that the summation is within the bounds necessary for classification.
\end{enumerate}


To gain as much performance boost as possible out of the individual experts, we adopt the third strategy outlined. This form enables the network to adapt the weights based on the input, allowing it to selectively tune the contribution of each expert based on their learned strengths and weaknesses.

For efficient use of the experts, it is necessary that the output of each expert is sufficiently different. Inherent in naming the individual algorithms "experts", this requires a certain level of specialization amongst them.  As \cite{ensembleDiversity} eloquently puts it, due to the inherent generalization errors produced by each expert, the collective opinions may overcome such errors and be more capable of generalization than any one network.

\section{Ensemble Convolutional Neural Networks}
\label{sec:moe_grasping}

We present the MOE model as applied to the problem of grasp accuracy prediction.  These models are called Ensemble Convolutional Neural Networks (ECNN).  ECNNs contain a set of grasping experts and a gating neural network responsible for combining the outputs of the experts to obtain a more accurate grasp quality estimation.

In an ECNN, we have a set of grasping experts $y_i$.  Each expert takes in sensory input $x$, which is a combination of an image of the object to be grasped $\mathbf{I}$, and the grasp position on the provided image $(u, v, d, w, \theta)$.  The output of each expert is an estimated grasp quality $\tilde{q}_i$.  The goal of the ECNN is to combine the grasp qualities into something closer to ground truth $q$.

To accomplish the required quality combination, we employ a gating convolutional neural network, which also takes in as input $x$. The role of this gating network is to determine the weights $g_i(x)$ assigned to each expert, based on the input $x$. To accomplish such a task, we specify two different methods for providing the input to the gating network, and discuss what each method semantically means, and how it could affect the accuracy of the ECNN.

The first method involves sending as input to the gating network the image of the object to be grasped, $\mathbf{I}$.  This method is referred to as \textbf{ImECNN} (Image-dependant ECNN).  This leads to the weight function being calculated as

\begin{equation}
    g(x) = g(\mathbf{I})
\end{equation}

The gating network may then assign weights appropriate to each individual expert based on the image received and their proficiency at classifying the grasps on the object presented.  This method enables the algorithm to perform expert discrimination on a per-image basis.

The second method involves integrating the targeted grasp into the gating network's input.  We call this network \textbf{GrImECNN} (Grasp-Image-dependant ECNN).  This leads to the weight function being calculated as

\begin{equation}
    g(x) = g[\mathbf{I}, (u, v, \theta)]
\end{equation}

To perform such a task, we crop the image of the object about the center of the grasp $(u, v)$ to be performed, and rotate the image by $\theta$ to align with the grasp orientation.  The crop amount determines the amount of emphasis we place on local information for the sake of classification.  In our experiments, cropping the image by 64×64 to 128×128 pixels seems to be the most performant.  The results presented for both the Cornell Accuracy results, and the experimental sections, have all GrImECNN gating networks trained on 64x64 cropped images.  This enables the gating network to perform expert discrimination on a per-object-per-grasp basis, allowing the use of different weights within the same image.  This approach introduces the greatest flexibility in terms of weight assignment, and can enable a more fine-grained approach by taking into account local and grasp-specific information.

In addition to the method of grasp expert separation, the decision must also be made on what the input and output of the ECNN must be.  In order to ensure compatibility with what the individual experts may require as I/O, and to ensure their outputs are properly combined, we define Data Adapters that can enable a modular, flexible means of integrating different grasping experts into the ECNN, as explained in the next section.

\section{Implementation}
\label{sec:implementation}
We define special meta preprocessing layers, which we call I/O Data Adapters, to transform the inputs and outputs of each grasping expert into something that can be combined by the full network. This maintains generality by removing the assumption of grasping expert homogeneity, especially with the potentially large I/O differences between experts.

Let the input to the entire model be $x$; in our case it would be the object image $\mathbf{I}$.  Since the input to the MOE should be sent to each expert, the common input must be transformed into an expert-usable format on a per expert basis.  The input to each expert is thus a transformation of the input.  We therefore now define the Input Data Adapter for expert $i$ as

\begin{equation}
    \tilde{x}_i = \operatorname{IDA}_i(x_i)
\end{equation}

The adapted input to each expert is $\tilde{x}_i$.  The output, to be combined, must have the same format as the expected output of the MOE as a whole.  Therefore, we must transform each expert's output into this common format.  Defining the output of expert $i$ as $\tilde{e}_i$, we define the Output Data Adapter as

\begin{equation}
    e_i = \operatorname{ODA}_i(\tilde{e}_i)
\end{equation}

\begin{figure}
  \includegraphics[width=3.0in]{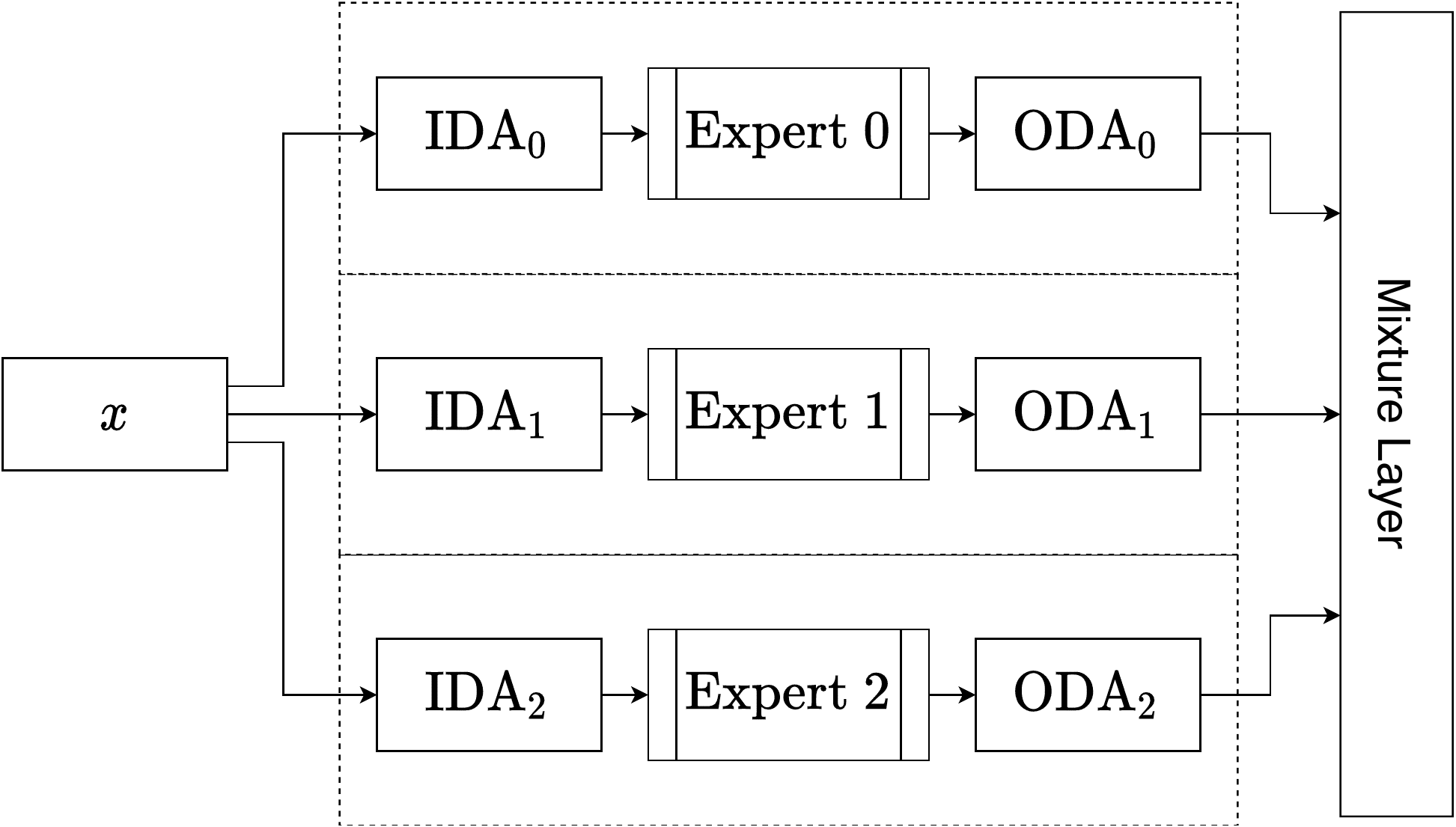}
  \caption{A network augmented with data adapters, enabling the flexible, modular use of highly different experts}
  \label{fig:dataAdapters}
 \vspace{-12pt}
\end{figure}

The implementation of the algorithm can be done in such a way that we may introduce minimal overhead and performance impacts.  Due to the disconnected nature of the gating network and the individual experts, the computer implementation of this problem is trivially parallelizable. All networks can be run in parallel, since each expert provides its output independently, while the gating network assigns the weights, also independently. The weighted outputs are then combined.  Therefore, the theoretical speed of the algorithm is limited by the speed of the slowest network in the MOE, assuming the networks can be run in parallel with no loss of performance or memory bottlenecks.

The training of an MOE model can be done efficiently as well by taking advantage of the frozen expert networks.  Since the gating network is the only one being trained, the outputs of the individual experts can be cached on disk beforehand, requiring only one run through over the data for each expert. Therefore, once trained, the experts can be evaluated on the training dataset, and the results cached and re-used.

\section{Training and Experimental Validation}
\label{sec:experimental_setup}

To verify the viability of the MOE model, we create a set of three ECNN networks, each utilizing a different method of opinion combination to compare and contrast the performance of each approach, and we train all three ECNNs on the Cornell Dataset.  The input to this network is an RGB-D image of the object to be grasped, along with a grasp position $(u, v, d, w, \theta)$.  The output is the estimated quality of the grasp.

The first expert to be used is GQCNN-4.0. This network was trained on the Dexnet Synthetic Dataset, a large dataset of objects with labeled grasps based on a robust analytical quality metric. GQCNN takes as input a depth image of the object, and the grasp position, and returns the quality of the given grasp.

The second expert is GGCNN. This expert was trained on the Cornell Dataset, which contains labeled grasping rectangles of object images taken with a real camera.  Hence its input has a different form compared to form of GQCNN's inputs. The input of this network is a depth image, but it is generative in that it associates a grasp quality, angle, and depth to each pixel in the image in one shot.

The final expert is a custom-trained version of GGCNN.  This custom version takes RGB images instead of depth images, but is identical otherwise. This provides us with an expert that takes in a completely different set of image channels compared to the previous two.

Because of the I/O difference in inputs and outputs GGCNN and GQCNN-4.0 expect, we need to resolve the variable I/O requirements and add all the relevant I/O Data Adapters.  Our full network input is a 300×300 RGB-D image $\mathbf{I}=\mathbb{R}^{300 \times 300 \times 4}$, and a grasp position $(u, v, d, w, \theta)$.  The adapters for each network where chosen such that we conform to the required I/O of each expert. For each expert, the input data adapters strip away unused image channels. The grasp data for GGCNN is discarded by its input adapter, as that is unused. The output adapters likewise convert the output of each expert into a scalar quality. For GQCNN-4.0, the output is unchanged. For both GGCNNs, the output data adapter retrieves the quality at the required grasp location, while for GQCNN-4.0 the output data adapter performs no transformation.

We implement three different ECNNs using the experts. We compare their relative performance to showcase the potential gains to be obtained with a more flexible weight distribution function. The first ensemble to be trained is the point of comparison, being a ECNN where the combination is done by a set of learned weights independent of the input. The next version is the ImECNN model, with the third being GrImECNN.

\begin{figure}
  \includegraphics[width=3.5in]{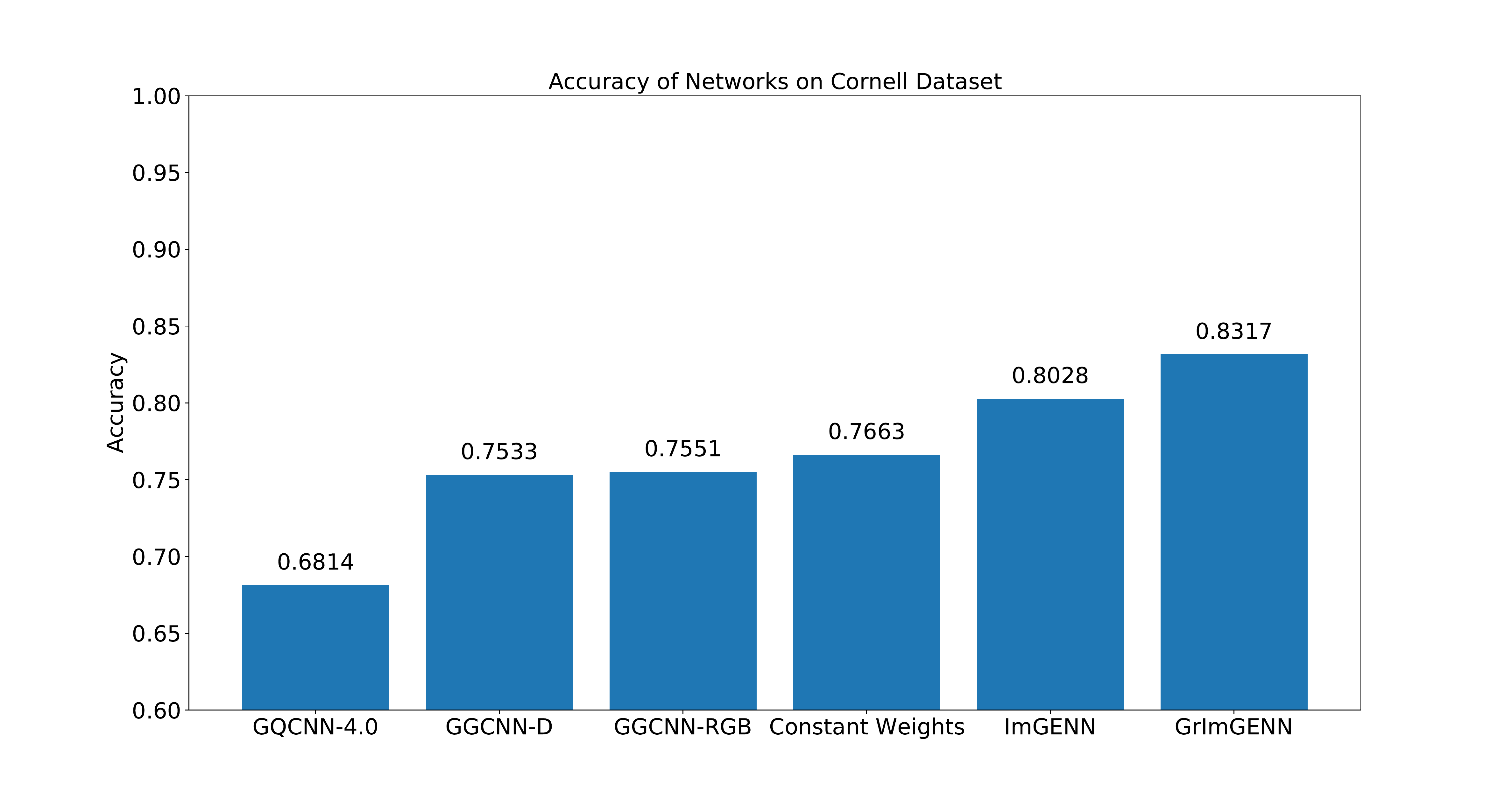}
  \caption{Comparison of the achieved accuracies on the Cornell Dataset for each of the networks.}
  \label{fig:comp}
 \vspace{-12pt}
\end{figure}

The training of the gating networks is done on the Cornell Dataset. After a fine-tuning training step to make GQCNN-4.0 more compatible with the dataset, we evaluate each expert on the entire dataset and cache the grasps on disk. The three separate ensemble models are implemented and trained in Keras using those cached results to simulate expert opinion using binary cross-entropy loss.

To evaluate the relative performance of each network, we compare the accuracy of each expert and ECNN in Fig.~\ref{fig:comp} in grasp classification. Here we define accuracy as the number of correctly labelled grasps (success or failure) over the total number of grasps.  Both GGCNNs were evaluated using their built-in evaluation algorithms, which are based on the intersection over union criteria, suitable for generative-based grasp synthesis algorithms (this agrees with outside measurements of its accuracy, e.g. \href{https://paperswithcode.com/paper/closing-the-loop-for-robotic-grasping-a-real}{\texttt{PapersWithCode}}). GQCNN-4.0, due to the difference in datasets, had to be fine-tuned using its built in fine-tuning functions, which explains the lowered relative accuracy, as compared to its performance in the physical experiment. A converter was written to transform the Cornell Dataset from a set of point clouds and grasp rectangles to depth images and grasp quality metrics stored in GQCNN-4.0's TensorDataset object, defined by the \texttt{autolab\_perception} package.

As we can see from Fig.~\ref{fig:comp}, there is a significant accuracy increase as we improve the ability of the gating network to make more fine-grained decisions based on the input, enabling it to select appropriate expert weights on an object-grasp basis.

The GrImECNN implementation of the previous set of ECNNs was used in a physical experiment on a robot to evaluate its performance in a typical grasping scenario. The experiment was run on a Franka Emika 7-DOF robot. The objects used are a set of 10 objects taken from the YCB Dataset\cite{ycb_dataset}.  Objects were placed in 3 random poses on a table at ground level in front of the robot. The robot was instructed to move to an overhead position and take an RGB-D picture using an Intel RealSense camera. Once all networks have resolved their calculations given the input, the combination happens with the weights obtained from the gating network. The grasp with the highest combined score is then attempted, and a success/fail is recorded based on whether the robot can lift the object. The outcome of the experiment can be seen in Table \ref{table:results}. Note that the numbers in bold correspond to the best performing algorithms for that YCB object.

\begin{table}[]
\begin{center}
\caption{Experimental results comparing the performance of the ensemble with the individual experts.}
\label{table:results}
\begin{tabular}{l|l|l|l|l}
\hline
\multicolumn{1}{c|}{\multirow{2}{*}{\textbf{Objects}}} & \multicolumn{4}{c}{\textbf{Success}}                                            \\ \cline{2-5} 
\multicolumn{1}{c|}{}                                   & GQCNN & Gen-RGB & Gen-D & GrImECNN\\ \hline
Screwdriver                                             & \textbf{3/3}     & 2/3                    & \textbf{3/3}           & \textbf{3/3}\\
Windex                                                  & 2/3              & 2/3                    & \textbf{3/3}        & 2/3\\
Mustard                                                 & 1/3              & 1/3                    & \textbf{3/3}        & \textbf{3/3}      \\
Bleach                                                  & 2/3              & 2/3                    & 2/3                 & \textbf{3/3}           \\
Pear                                                    & \textbf{2/3}     & 0/3                    & 1/3                 & \textbf{2/3}       \\
Banana                                                  & \textbf{3/3}     & \textbf{3/3}           & \textbf{3/3}        & \textbf{3/3}  \\
Mug                                                     & 1/3              & 1/3                    & \textbf{2/3}        & \textbf{2/3} \\
Spatula                                                 & \textbf{3/3}     & \textbf{3/3}           & 2/3                 & \textbf{3/3} \\
Spring Clamp                                           & \textbf{3/3}     & \textbf{3/3}           & 2/3                 & \textbf{3/3} \\
Wine Glass                                              & 0/3              & 0/3                    & \textbf{1/3}        & \textbf{1/3} \\ \hline
Total                                                  & 20/30              & 17/30                   & 22/30                & \textbf{25/30} \\ \hline
\end{tabular}
\end{center}
\end{table}

As can be seen from Table \ref{table:results}, the adopted GrImECNN model is the best performer, with an accuracy that is often equal to, or in some cases surpassing, the best performing expert in the MOE. This showcases model's ability to leverage the strengths of each expert to improve overall system accuracy. It achieved only 5 losses out of a total of 30 grasping attempts. On the Bleach bottle, the GrImECNN managed to find a combination of experts that outperformed the rest, resulting in an increase in accuracy compared to the best-performing singular expert.  The GrImECNN managed to find, based on the image and grasp being attempted, an optimal combination of the expert opinions such that accuracy was maximized.

\section{Conclusion}
\label{sec:conclusion}

We have showcased throughout this paper how to combine a set of diverse grasping experts in a modular fashion within the MOE framework. By presenting two different approaches for expert classification based on input, we can choose which parameters affect weight assignment. We illustrated the beneficial nature of such a combination and how it enhances the performance of the entire grasping system by taking advantage of the strengths and specialties of each grasping expert.  Guidelines for training using the disk caching of expert outputs, and for implementation using parallelism were presented as well, all used to reduce the overhead and impact of implementing and training the more complex MOE framework.  In addition, although the results and networks presented here work on single-grasp quality estimation, a generative gating network can be used with generative experts instead to enable a full generative ensemble, which may provide better overall performance and accuracy.

We believe that the strategy of benefiting from different experts opinions can be transformative for the robotic grasping field. We plan to explore alternative ways of expert combination, and to examine what combination of experts may produce the best combined outcome in terms of both accuracy and performance.

\bibliographystyle{IEEEtran}
\bibliography{root.bib}

\end{document}